\documentclass[letterpaper, 10 pt, conference]{ieeeconf}  

\IEEEoverridecommandlockouts                              

\usepackage{color}
\usepackage{graphics} 
\usepackage{epsfig} 
\usepackage{mathptmx} 
\usepackage{times} 
\usepackage{amsmath} 
\usepackage{amssymb}  

\makeatletter
\newcommand*\titleheader[1]{\gdef\@titleheader{#1}}
\AtBeginDocument{%
  \let\st@red@title\@title
  \def\@title{%
    \bgroup\normalfont\small\flushleft\@titleheader\par\egroup
    \vskip.5em\st@red@title}
}
\makeatother

\title{\LARGE \bf Accurate Cooperative Sensor Fusion by Parameterized Covariance Generation for Sensing and Localization Pipelines in CAVs}

\titleheader{2022 IEEE Intelligent Transportation Systems Conference (ITSC)}

\author{Edward Andert and Aviral Shrivastava
\\
Arizona State University%
\thanks{This work is partially supported by National Science Foundation under Grant Numbers CPS 1645578 and CCF 1723476 and the NSF/Intel joint research center for Computer Assisted Programming for Heterogeneous Architectures (CAPA).}
}

\begin{document}

\maketitle
\thispagestyle{empty}
\pagestyle{empty}

\begin{abstract}

A major challenge in cooperative sensing is to weight the measurements taken from the various sources to get an accurate result. Ideally, the weights should be inversely proportional to the error in the sensing information. However, previous cooperative sensor fusion approaches for autonomous vehicles use a fixed error model, in which the covariance of a sensor and its recognizer pipeline is just the mean of the measured covariance for all sensing scenarios. The approach proposed in this paper estimates error using key predictor terms that have high correlation with sensing and localization accuracy for accurate covariance estimation of each sensor observation. We adopt a tiered fusion model consisting of local and global sensor fusion steps. At the local fusion level, we add in a covariance generation stage using the error model for each sensor and the measured distance to generate the expected covariance matrix for each observation. At the global sensor fusion stage we add an additional stage to generate the localization covariance matrix from the key predictor term velocity and combines that with the covariance generated from the local fusion for accurate cooperative sensing. To showcase our method, we built a set of 1/10 scale model autonomous vehicles with scale accurate sensing capabilities and classified the error characteristics against a motion capture system. Results show an average and max improvement in RMSE when detecting vehicle positions of 1.42x and 1.78x respectively in a four-vehicle cooperative fusion scenario when using our error model versus a typical fixed error model.

\end{abstract}

\section{Introduction}


Cooperative sensing has been proposed to mitigate sensor coverage and obstruction issues in autonomous vehicles. Cooperative sensing occurs when multiple connected autonomous vehicles (CAVs) combine their data together to get a more accurate picture of the world around each individual CAV \cite{feng2020deep}. Cooperative sensor fusion has been proposed to improve a number of systems in autonomous vehicle including localization \cite{li2012multi, bresson2016cooperative, de2016cooperative} and perception \cite{chen2019cooper, arnold2020cooperative}. Additional connected infrastructure sensors (CISs) that are placed throughout the city (such as traffic cameras) could be used to gather more data for the cooperative sensors fusion and strengthen the robustness \cite{arnold2020cooperative}. 

Most prior works on cooperative fusion consider a fixed error model, e.g. creating a covariance matrix for sensor error using the mean error seen in all scenarios. However, assuming sensor errors to be fixed can result in poor weighting of observations in cooperative fusion. For instance, let us consider a scenario where two identical cameras report the distance to an object. Sensor A is 5 meters away from the object and sensor B is 200 meters away. Clearly we should know that the closer camera should be weighted more heavily. However, a fixed error model will not consider this problem and will weight the data the same. Consider a second scenario where vehicles A and B have a localizer that works well when traveling slow but badly at a faster speed due to a slow update frequency. The two vehicles are equidistant from an object and have the same sensor suite, but vehicle A is traveling 10x slower than vehicle B. It seems clear that the result from vehicle A should be weighted much higher than B because B will be reporting the position of the object with respect to its localization data which is less accurate. In these two scenarios a fixed error model would weight both sensed values the same and the cooperative result may be worse than the ego vehicle sensing alone. Therefore, changes in both perception error and localization error must be taken into account when performing cooperative sensing.

In this paper, we: 
\begin{itemize}
\item Analyze the error sources in autonomous vehicles and pinpoint that distance from the sensor is a good predictor of sensing error, and that the velocity is a good predictor of the localization error. Using these predictors enables the generation of more accurate covariance estimation for each sensor observation.
\item Add a parameterized covariance generation step to the local fusion process based on the sensor pipeline characterization that uses distance as a predictor to get a better covariance estimate. Add a parameterized localization covariance generation using velocity as a predictor, and combine it with the local sensor fusion result to drive the global fusion step. This results in a more accurate cooperative sensing.
\end{itemize}

To demonstrate and evaluate the effectiveness of our approach, we perform an in depth analysis of our parameterized error model on a 1/10 scale autonomous vehicle setup consisting of up to four CAVs and two CISs using a motion capture system as a baseline. Results from our work show a significantly improvement in error fitment using our parameterized error model vs. fixed error model on on our 1/10 scale setup. We run a high level sensor fusion pipeline with Joint Probability Data Association Filter (JPDA) and Extended Kalman Filter (EKF) to match and perform cooperative sensor fusion on our 1/10 scale setup using various scenario settings. The results of these tests show our parameterized error model to be 1.42x more accurate in terms of RMSE versus the motion capture system baseline across our test set and has up to a 1.78x improvement for the best case.

\section{Related Works}
Stroupe \textit{et al.} proposed a covariance estimation technique that used measured distance as a predictor for the $<x,y>$ measurement covariance of a football detected by their football-playing robots \cite{stroupe2001distributed}. Though limited to a camera sensor modality, they showed a significant correlation with error vs. distance and that measured distance could be used as a predictor. We have adopted the term ``parameterized error model'' for generating a covariance matrix using a predictor terms (like distance by Stroupe \textit{et al.}) to get a better fitment to the error data whereas the typical approach is to use a ``fixed error model'' where the covariance of a sensor and its recognizer is generated using the mean error for all sensing scenarios. This parameterized error estimation seems to have not crossed over to the autonomous vehicle field from the multi-robot field even through autonomous vehicles sensor have been shown to exhibit the same relationship. Garcia \textit{et al.} showed a significant correlation between LIDAR sensor distance measurement error with the distance to that object supporting the need for a parameterized error model \cite{garcia2012environment}. Chadwick \textit{et al.} go on to show the same distance versus measurement error relationship for both camera and radar detecting vehicles and how it affects Recall as well \cite{chadwick2019distant}. A lone AV is likely to not notice the distance accuracy relationship as it's own sensors see the same objects from a similar distance due to being mounted on the same rigid body so the problem is masked. However, this distance error relationship is very important for cooperative sensing because of the drastically different distances sensors can observe the same object from.

A further problem in cooperative fusion involving autonomous vehicles is localization error. Wan \textit{et al} showed that localization error of multiple types of localizers vs. ground truth is different on the longitudinal axis of a vehicle than on the lateral axis and can be modeled using an ellipse \cite{wan2018robust}. From the perspective a singular CAV, this localization error and its directionality is not critical. However, when we move to a cooperative sensing this localization has an additive effect because it is passed along to all observation from the CAV and is thus essential to model correctly.

Prior approaches to the problem of cooperative sensor fusion in autonomous vehicles tend to take one of three approaches for sharing data: 1) sending raw sensor data \cite{kim2013cooperative, kim2014multivehicle, chen2019cooper} which is known as early fusion or low level sensor fusion, 2) sending object position with relevant type information and bounding box which is known as late or high level sensor fusion \cite{aeberhard2011high, arnold2020cooperative, allig2020unequal}, and finally, 3) sending Voxel occupancy grids \cite{chen2019f} which we define as a hybrid fusion approach. We eliminate approach one due to well cited communication delay and scalability issues even using 802.11n in close proximity \cite{kim2013cooperative}. We also eliminate the hybrid approach three due to scalability problems with sending Voxel occupancy grids along with a severe accuracy vs. performance trade-offs that must be decided when choosing the size of the Voxels \cite{aeberhard2011high}. We chose approach 2, high level fusion, because it does not suffer from communication issues and it allows for a more intuitive approach for linking sensor accuracy and modality to their respective observations \cite{feng2020deep}.

Rauch \textit{et al.} proposed a high level cooperative fusion approach using road side units and other vehicles to improve localization of an ego vehicle and compared this result to a DGPS baseline \cite{rauch2012car2x}. Their method coined the terms ``local fusion'' and ``global fusion'' to refer to the fusion performed on sensors on-board an ego vehicle and the fusion performed on inputs from all sensors in the area respectively, both terms we adopt in this paper. However, their results were limited to improving localization and not perception. Arnold \textit{et al.} compare low level and high level cooperative fusion techniques using CIS to augment the CAVs but their results do not give statistics on measurement variance and instead focus on recall in addition to using a fixed sensor error model\cite{arnold2020cooperative}. Tsukada \textit{et al.} published a paper and article respectively proposing high level V2X fusion method and software called C2X and provide an analysis of the results at scale using SUMO and shows the viability of high level cooperative V2X fusion at scale\cite{tsukada2020autoc2x, tsukada2020networked}. However, the authors use a fixed sensor error model and did not account for localization covariance. Finally, Allig \textit{et al.} proposed a method for fusing heterogeneous sensors to track, using a covariance intersection method that is very similar to the fusion pipeline we use in this paper \cite{allig2020unequal}. Again, the authors use a fixed error model for the generation of sensor measurement and did not account for localization covariance.

\section{Experimental Setup}
The first contribution of this paper is to analyze and characterize sources of error within autonomous vehicles to look for trends that can be exploited for cooperative fusion. Therefore we first discuss our experimental setup before moving on to our findings and our results. We develop a 1/10 scale testbed with 1/10 scale vehicles, 1/10 scale road dimensions, 1/10 scale object dimensions, etc. This setup has the advantage that we can utilize an Optitrack System to collect the ground truth.

\begin{figure}[ht]
  \centering
  \begin{minipage}[b]{0.23\textwidth}    \includegraphics[width=\textwidth]{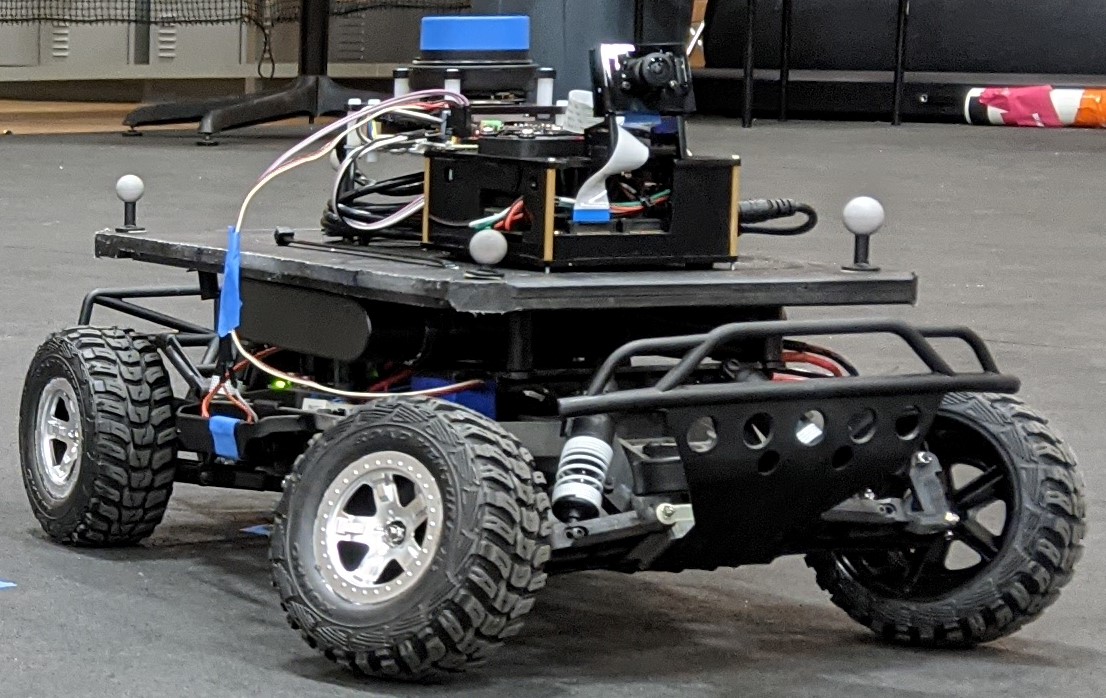}
    \caption{One tenth scale CAV with camera, LIDAR, and Nvidia Jetson Nano for on-board processing.}
    \label{fig:one_tenth_cav}
  \end{minipage}
  \hfill
  \begin{minipage}[b]{0.23\textwidth}
    \includegraphics[width=\textwidth]{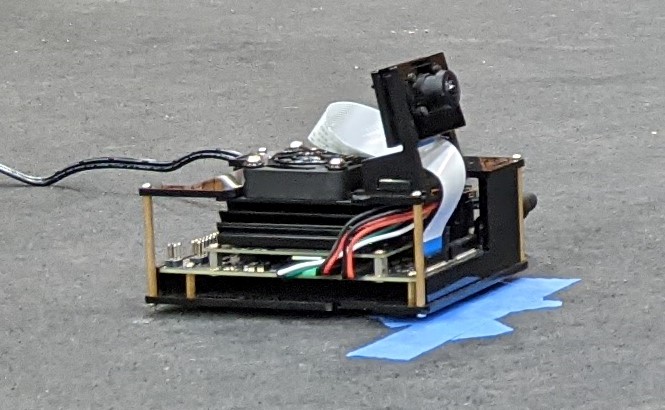}
    \caption{One tenth scale CIS traffic camera replica using Nvidia Jetson Nano and IMX160 camera in a case.}
    \label{fig:one_tenth_cam}
  \end{minipage}
\end{figure}

\subsection{1/10 Scale Connected Autonomous Vehicle (CAV)}
We outfitted four 1/10 scale vehicles with scale accurate autonomous vehicles sensors including a LIDAR and camera. Sensor data is processed on-board the 1/10 scale vehicle using an Nvidia Jetson Nano 4GB. Hardware is mounted on a Traxxis Slash 1/10 scale RWD remote control vehicle with controls of the motor and steering servo performed using a PCA9685 board connected to the Jetson Nano.

\subsection{1/10 Scale Connected Infrastructure Sensor (CIS)}
We created 1/10 scale traffic cameras using Nvidia Jetson Nano and IMX160 camera used on the vehicles. The main difference in this setup from the vehicle setup is the lack of LIDAR and that fact it is a stationary platform.

\begin{table}[]
\caption{\textbf{1/10 Scale CAV Sensor Hardware}}
\label{table:cav_hardware}
\begin{tabular}{lllll}
 \textbf{Type} & \textbf{Model} & \textbf{Bearing} & \textbf{FOV} & \textbf{Pipeline} \\
 LIDAR & Slamware M1M1 & 0$^{\circ}$ & 360$^{\circ}$ & SLAM, DBScan \\
 Camera & IMX160 & 0$^{\circ}$ & 160$^{\circ}$ & YoloV4 CNN \\
\end{tabular}
\end{table}

\begin{table}[]
\caption{\textbf{1/10 Scale CIS Sensor Hardware}}
\label{table:cis_hardware}
\begin{tabular}{lllll}
 \textbf{Type} & \textbf{Model} & \textbf{Bearing} & \textbf{FOV} & \textbf{Pipeline} \\
 Camera & IMX160 & 0$^{\circ}$ & 160$^{\circ}$ & YoloV4 CNN \\
\end{tabular}
\end{table}

\subsection{Camera Object Detection Pipeline}
The IMX160 camera recognition pipeline consists of YoloV4 Tiny convolutional neural net (CNN) running natively on the Jetson Nano GPU. The CNN has been trained to recognize other 1/10 scale cars, small 1/10 scale cones, and lane corners. Distance to the object recognized by the camera is estimated using the known height of the object model, focal length of the camera, and the height detected in pixels which allows us to compute the distance of the detected object. We run the camera recognition pipeline at 8 Hz to match the LIDAR.

\subsection{LIDAR Object Detection Pipeline}
The LIDAR recognition pipeline consists of gathering the angle and distance measurements from the LIDAR, converting to a point cloud, and using the DBScan library to cluster the points. We then fit the shape of the object to determine the distance and angle to the centroid of the vehicle point cluster.

\subsection{Localization Pipeline}
Localization is performed via SLAM on-board the Slamtec M1M1 LIDAR along with its integrated IMU and outputted at 8 Hz. This process is proprietary though it is known to be SLAM and works up to a velocity of 1 meter per second.

\subsection{Driving Scenarios}
\begin{figure}[ht]
  \centering
  \begin{minipage}[b]{0.23\textwidth}    \includegraphics[width=\textwidth]{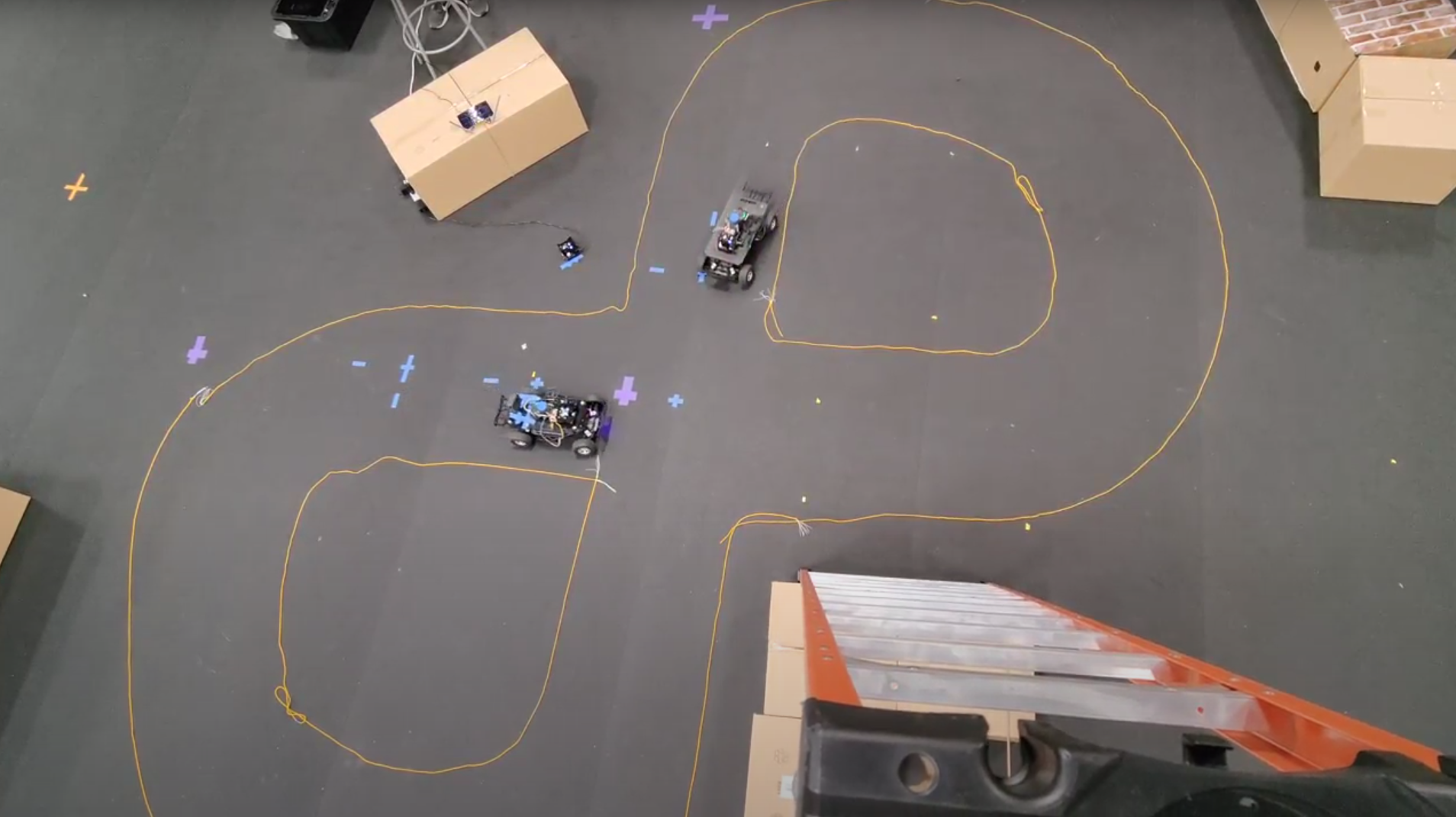}
    \caption{Small ($s_l = 1.0m$) figure 8 loop with two CAVs and one CIS. Cardboard boxes around the track were placed for localization.}
    \label{fig:top_down}
  \end{minipage}
  \hfill
  \begin{minipage}[b]{0.2\textwidth}
    \includegraphics[width=\textwidth]{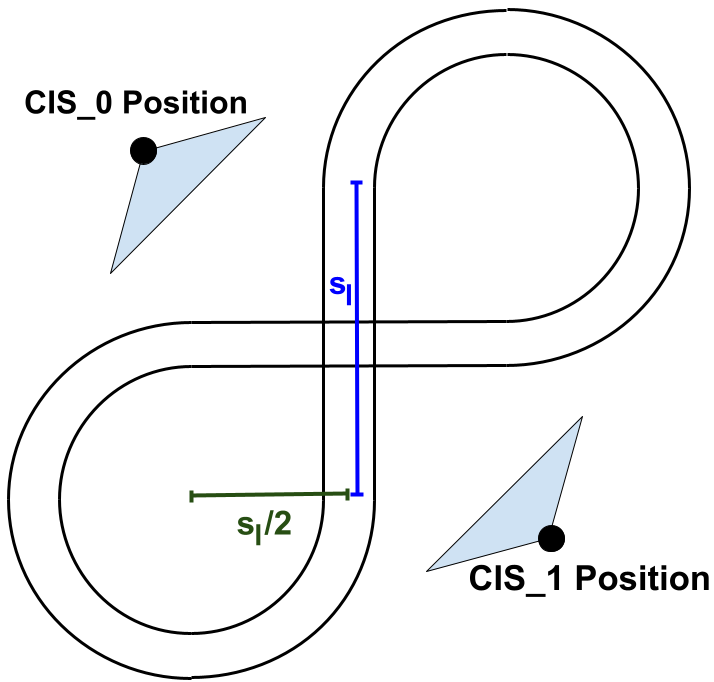}
    \caption{Generalized format for figure 8 loop showing straight length $s_l$. Positions for CIS sensors are indicated.}
    \label{fig:cav_locaalization_latlon}
  \end{minipage}
\end{figure}

Our 1/10 vehicle setup utilizes a figure 8 route. The figure 8 is the optimal route for testing CAVs as it supplies an intersection, some straight driving, and two turns that are in the opposite direction such that there is no turn directional bias. The figure 8 is defined by the straight length, which we will call $s_l$. The turns are of constant radius which is equal to $s_l/2$. We have created two different figure 8 tracks, one where $s_l = 1.0m$ and the other where $s_l = 2.0m$. This gives us both a small and large map to test the effect of distance on sensors.

We vary the amount of vehicles and sensors in the figure 8 setup. We create two scenarios, a low density scenarios where there is a CIS and two CAVs, and a high density scenario where there are two CISs and four CAVs. Utilizing the two track setups and two density scenarios, we get four total test setups. We then further vary whether CIS sensors are present or not as depicted in table \ref{table:scenarios} for a total of eight scenarios. We believe these eight scenarios give reasonable coverage of the main scenarios that CAVs would encounter, light density small area, light density large area, high-density small area, and a high-density large area along with whether there are CIS sensors installed or not.

\begin{table}[]
\caption{\textbf{1/10 Scale Vehicle Figure 8 Test Scenarios}}
\label{table:maps}
\begin{tabular}{lllll}
 \textbf{Name} & \textbf{Explanation} & \textbf{$s_l$} & \textbf{CAVs} & \textbf{CISs} \\
 sm/sp & \textit{small, sparse, no CIS} & 1m & 2 & 0 \\
 sm/de & \textit{small, dense, no CIS} & 1m & 4 & 0 \\
 lg/sp & \textit{large, sparse, no CIS} & 2m & 2 & 0 \\
 lg/de & \textit{large, dense, no CIS} & 2m & 4 & 0 \\
 sm/sp/CIS & \textit{small, sparse, with CIS} & 1m & 2 & 1 \\
 sm/de/CIS & \textit{small, dense, with CIS} & 1m & 4 & 2 \\
 lg/sp/CIS & \textit{large, sparse, with CIS} & 2m & 2 & 1 \\
 lg/de/CIS & \textit{large, dense, with CIS} & 2m & 4 & 2 \\
 \label{table:scenarios}
\end{tabular}
\vspace{-0.2in}
\end{table}

Error classification of the 1/10 scale vehicles is relatively straightforward. We outfit all of the CAVs with Optitrack motion capture trackers. The CISs and CAVs are placed in predetermined positions. We zero out the offsets between the Optitrack and Slamtec M1M1 coordinate systems before each test. Each scenario is run for 10 minutes with the Optitrack system recording ground truth data at 120Hz. Time synchronization is periodically performed between the Optitrack and 1/10 scale system. Data from each vehicle, sensor, and the Optitrack system is saved so it can be parsed and replayed to check for specific parameters.

\section{Error Classification Results}
\subsection{Velocity as a Localization Predictor}
Localization accuracy is the first error classification that we performed. At a larger scale, this data could be latitude, longitude, and heading. In our small setup, this is a more simple x, y, yaw where the center of figure 8 is $x = 0$ and $y = 0$. Localization accuracy is measured against the motion capture system using the scenarios depicted previously.

\begin{figure}[ht]
    \includegraphics[width=.45\textwidth]{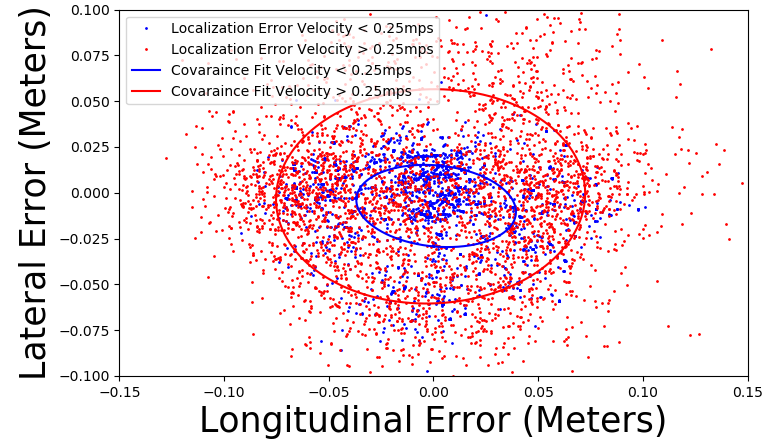}
    \caption{Longitudinal and lateral localization error components scatter plot W.R.T. the vehicle heading. Red points indicate a velocity between .25 and .5 meters per second and blue indicates a velocity between .25 and 0 meters per second. This plot shows a covariance fit to both of these data-sets drawn as an ellipse in the same color. One can clearly see a significant correlation between velocity and localization error. RMSE is 0.0576m}
    \label{fig:cav_locaalization_xy}
\end{figure}

We graphed the localization error with respect to the ego vehicle direction of travel laterally (side to side) and longitudinally (front to back) in figure \ref{fig:cav_locaalization_xy}. We color the points according to the velocity, blue to indicate a velocity less than .25 meters per second, and red to indicate the velocity is greater than .25 meters per second. We fit both sets of data with a 2x2 covariance matrix using the least squares method. This results in the two ellipses in figure \ref{fig:cav_locaalization_xy}, the blue ellipse is clustered tightly around the origin and the red ellipse is far more spread out. These different ellipses can be explained from the stop-and-go nature of the figure 8 path we drive. Our vehicle is either maintaining the target speed, .5 meters per second, or is stopped or slowing for the red traffic light and the velocity is near 0. Our hypothesis that the error was Gaussian was rejected according to D’Agostino’s $K^2$ test, however we are still able to prove the premise that velocity is a useful predictor of variance. A linear regression fitment to the data using velocity as the X axis and expected error as the Y axis results in an $R^2$ fitment of .22 longitudinal error and .18 for lateral error. This is not a great fit, but it is significantly better than the fixed average (mean) which has an $R^2$ fitment of 0. Therefore, our analysis of localization error of our 1/10 CAVs shows that we can predict the position variance based on measured velocity better than the mean.



\subsection{Distance as a Object Detection Predictor}
We define object detection as the ability to detect the centroid, bounding box, and type of an object using a sensing pipeline such as camera or LIDAR. For the purposes of this paper, we only consider the centroid position when determining object detection accuracy. To measure object detection accuracy, we recorded distance and angle to all objects tracked in the scene as reported by the LIDAR and camera detection pipelines separately. We use global nearest neighbors to match observation to ground truth from the Optitrack system, along with human labeling where required. We compare the distance and angle reported by the Optitrack to that reported by the sensor. The result is two reported characteristics, delta distance and delta angle. This removes localization as a bias by utilizing the true position as reported by Optitrack rather than the localizer report.

\begin{figure}[ht]
  \centering
    \includegraphics[width=.48\textwidth]{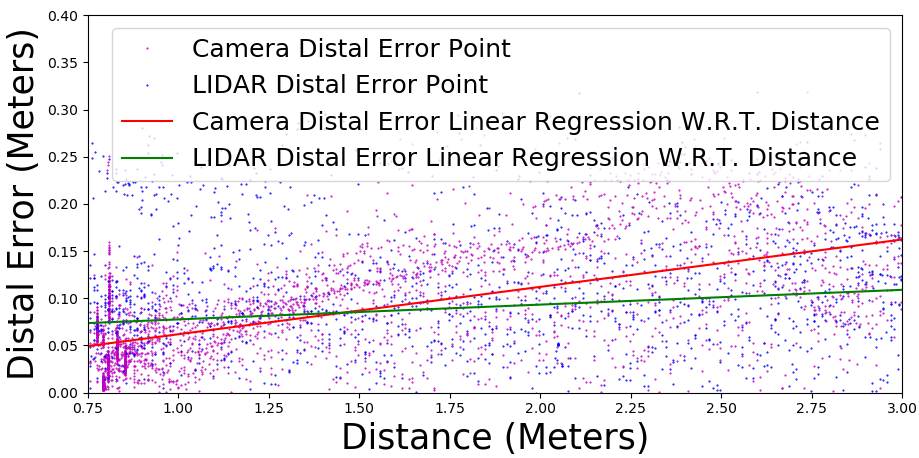}
    \caption{Plotting distal error of an object detection versus distance detected to that object results in a clear correlation. Camera and LIDAR linear regression $R^2$ fitness are 0.18 and 0.35 respectively.}
    \label{fig:cav_distal}
\end{figure}

Plotting the error in distance detected to the object versus detected distance to the object in figure \ref{fig:cav_distal} we get a strong correlation. There are some artifacts at short range which come from the vehicles sitting in the same stationary location when the traffic light is red as well as more noise for the LIDAR. We believe this LIDAR noise is because the LIDAR pipeline is much worse than the camera at determining what is a CAV or just a random object so it may be mis-detecting the corners of boxes as other CAVs resulting in a higher rate of outliers. LIDAR error does not increase as fast as camera error and there is a distinct crossover point where the LIDAR becomes more accurate than the camera at ~1.5 meters distance. We have not shown the perpendicular error here but the $R^2$ is 0.021 and 0.045 for LIDAR and camera perpendicular error respectively.

\section {Cooperative Sensing Pipeline Overview}
The approach we choose for the sensor fusion pipeline is tied to our decision to utilize high-level (late) sensor fusion. Sensors and their recognition pipelines are tied to a sensor package which is a collection of sensors mounted on the same rigid body. We consider two types of sensor packages, 1) a connected infrastructure sensor (CIS) which could be a traffic camera or other statically mounted sensor package, and 2) a connected autonomous vehicle (CAV) which is a typical autonomous vehicle with a suitable sensor suite. All sensing platforms are assumed to include communication hardware such that they can talk to other sensor packages and roadside units (RSU) within range. Due to bandwidth constraints as well as the need for the CAV to drive using its sensor output, our cooperative sensor fusion is done in a cascade approach. On-board each sensor package, each sensor and its respective recognition algorithm(s) processes a frame of data from the sensor and outputs a list of object observations which consists of a centroid $<x,y>$, bounding box, and type for each observation. Local sensor fusion is performed on all on-board sensor data. Results of this local sensor fusion are transmitted to a local RSU that aggregates and fuses it with the data received from other CAVs and CISs in the area, and then this cooperative sensing data is distributed back out the CAVs so it can be consumed as sensing input.

\section{Local Fusion Design}
\begin{figure}[ht]
  \centering
    \includegraphics[width=.45\textwidth]{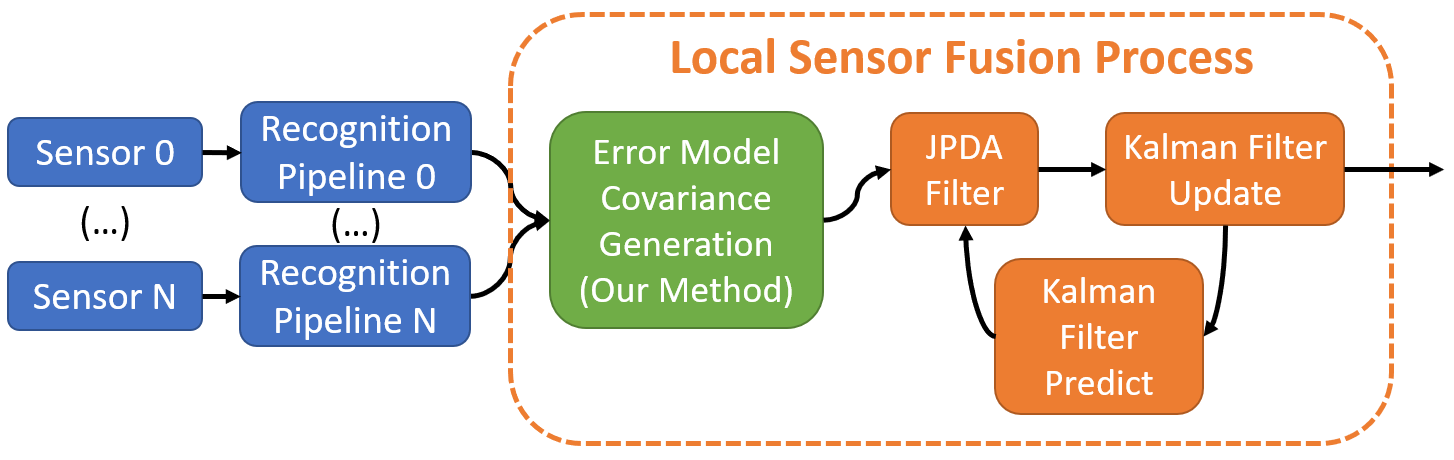}
    \caption{Our method for local sensor fusion adds in a covariance generation stage using the error model for each sensor and the measured distance to generate the expected covariance matrix for each observation.}
    \label{fig:local_fusion}
\end{figure}

Figure \ref{fig:local_fusion} depicts the local fusion process of a single CAV or CIS. Sensors along with their recognizer are treated as separate pipelines until the JPDA filter stage where all the observations are associated using the JPDA filter. Associated observations that are considered a single object are tracked using an EKF for that track. The EKF has a predict stage and then up to $n$ update stages depending on how many observations from the sensor pipelines are associated with the same track.

\subsection{Parameterized Sensing Error Model}

\begin{figure}[ht]
  \centering
    \includegraphics[width=.45\textwidth]{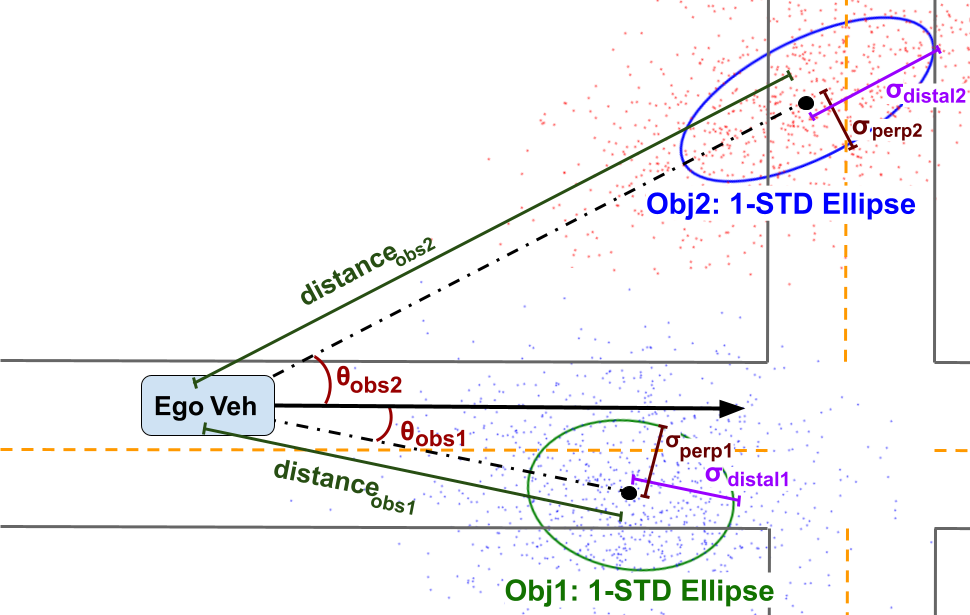}
    \caption{This figure depicts a scenario where a stationary ego vehicle (a 1/10 scale CAV) is sensing two stationary objects (other 1/10 scale CAVs) many times using a camera sensor mounted at $<0,0>$ with a facing angle of 0 degrees W.R.T. the ego vehicle. The resulting point distribution is shown with an estimated 2x2 covariance matrix shown as an ellipse that was generated by our camera error estimation. Note: the error distribution has been greatly exaggerated so that it is visible in this paper format.}
    \label{fig:explanaition_of_terms}
\end{figure}

In the prior section we observed that for the LIDAR and camera cases, measured distance as reported by the sensing pipeline was an accurate predictor of both distal and perpendicular error with respect to the sensor. We fit a prediction equation to the characterization error data. In our cases, we believe a linear regression fit is accurate enough, but if need be a polynomial fit could be used as well if it results in a better fit. Equation \ref{equation:measurement} shows the measurement data received from a sensor pipeline. Equation \ref{equation:distal} and equation \ref{equation:perpendicular} show the polynomial form for the distal and perpendicular error respectively where $\alpha_k$ and $\beta_k$ are the polynomial expansion coefficients that will be fit to the data using the measured distance $d$ to the observation as a predictor. Figure \ref{fig:explanaition_of_terms} shows the depiction of the terms used in Equations \ref{equation:measurement} through \ref{equation:perpendicular}.

\begin{equation}
obs = \big[\begin{smallmatrix}
  distance_{obs} \\
  \theta_{obs} 
\end{smallmatrix}\big]
\label{equation:measurement}
\end{equation}

\begin{equation}
\sigma_{distal} = \sum_{k=0}^{n} \alpha_k * distance_{obs}
\label{equation:distal}
\end{equation}

\begin{equation}
\sigma_{perp} = \sum_{k=0}^{n} \beta_k * distance_{obs}
\label{equation:perpendicular}
\end{equation}

Using equation \ref{equation:distal} and equation \ref{equation:perpendicular}, we generate the distal and perpendicular errors respectively with distance $d$ measured to the observed object as a predictor. The observation position $\mu$ is calculated from the sensor position $<x_{sensor}, y_{sensor}>$ and sensor angle $\theta_{sensor}$ with respect to the ego vehicle rear axle by transforming the observed object given in equation \ref{equation:measurement} to the ego vehicle coordinate system using equations \ref{equation:angle_transform} and \ref{equation:vehicle_transform}. We generate the covariance of the observed object $\Sigma_{obs}$ using the bearing angle to the object with respect to the rear axle or $\phi_{obs}$ and the expected distal error and expected perpendicular error given by equation \ref{equation:distal} and equation \ref{equation:perpendicular} respectively using equation \ref{equation:vehicle_transform} and \ref{equation:covariance_rotation}. This resulting $<x,y>$ location contained in $\mu_{obs}$ and 2x2 covariance contained in $\Sigma_{obs}$ are now in a form that can be easily fed into a fusion algorithm (Kalman Filter, EKF, UKF, etc.)

\begin{equation}
  \phi_{obs} = \theta_{sensor} + \theta_{obs} 
  \label{equation:angle_transform}
\end{equation}

\begin{equation}
\mu_{obs} = \big[\begin{smallmatrix}
  x_{sensor} + distance_{obs} \cos{\phi_{obs}} \\
  y_{sensor} + distance_{obs} \sin{\phi_{obs}}
\end{smallmatrix}\big]
\label{equation:vehicle_transform}
\end{equation}

\begin{equation}
\Sigma_{obs} = 
\big[\begin{smallmatrix}
  cos(\phi_{obs}) & sin(\phi_{obs}) \\
  -sin(\phi_{obs}) & cos(\phi_{obs})
\end{smallmatrix}\big]
\big[\begin{smallmatrix}
  \sigma_{distal} & 0 \\
  0 & \sigma_{perp}
\end{smallmatrix}\big]
\big[\begin{smallmatrix}
  cos(\phi_{obs}) & sin(\phi_{obs}) \\
  -sin(\phi_{obs}) & cos(\phi_{obs})
\end{smallmatrix}\big]^\intercal
\label{equation:covariance_rotation}
\end{equation}

\subsection{Local Joint Probability Data Association Filter Design}
We chose the Joint Probability Data Association (JPDA) Filter for the association of observations. Similar to García \textit{et al.}, each sensor observation is treated individually and is matched to the existing set of tracks \cite{garcia2017sensor}. Each track has an associated EKF for fusing the observations. If an observation is not in the gate of any track and is not considered noise, then a new track will be created. To smooth out some smaller errors, a track is not officially reported until it has been tracked for a minimum amount of frames. Conversely, if a track has not had an association from a sensor observation within a specified amount of frames, that track and its associated EKF are deleted. The covariance for each observation contained in $\Sigma_{obs}$ is fed into the JPDA Filter along with the observed location of the object, $\mu_{obs}$.

\subsection{Sensor Platform EKF Design}
All sensing pipelines are assumed to output an estimate for the $\mu_{obs}$ of each object they detect with respect to the sensor platform (e.g. CAV or CIS). The number of sensor pipelines attached to a sensor platform is not bounded. We chose to use the model introduced by Farag \textit{et al.}, with some minor modifications \cite{farag2021kalman}. Using the standard form for an EKF, $x_{k}$, $F_{k}$, $Q$, $z_{k}$, $R_{k}$, and $H_{k}$ are defined in equations \ref{equation:x_k_local}, \ref{equation:f_k_local}, \ref{equation:q_local}, \ref{equation:z_k_local}, \ref{equation:r_k_local}, and \ref{equation:h_k_local} respectively.


\begin{equation}
x_{k} = 
\begin{bmatrix}
  x \\
  y \\
  v \\
  \psi \\
  \dot{\psi} \\
\end{bmatrix}
\label{equation:x_k_local}
\end{equation}

\begin{equation}
\resizebox{\linewidth}{!}{$
F_{k} = 
\begin{bmatrix}
  1 & 0 & \frac{1}{\psi}(-\sin{\psi}+\sin{\Delta_t \dot{\psi} + \psi} & \frac{v}{\dot{\psi}}(-\cos{\psi}+\cos{\Delta_t \dot{\psi} + \psi} &
  \frac{v\Delta_t}{\dot{\psi}}(\cos{\Delta_t \dot{\psi} + \psi}) - \frac{v\Delta_t}{\psi}(-\sin{\psi}+\sin{\Delta_t \dot{\psi} + \psi} \\
  0 & 1 & \frac{1}{\psi}(\cos{\psi}-\cos{\Delta_t \dot{\psi} + \psi} & \frac{v}{\dot{\psi}}(-\sin{\psi}+\sin{\Delta_t \dot{\psi} + \psi} &
  \frac{v\Delta_t}{\dot{\psi}}(\sin{\Delta_t \dot{\psi} + \psi}) - \frac{v\Delta_t}{\psi}(\cos{\psi}-\cos{\Delta_t \dot{\psi} + \psi} \\
  0 & 0 & 1 & 0 & 0\\
  0 & 0 & 0 & 1 & \Delta_t \\
  0 & 0 & 0 & 0 & 1 \\
\end{bmatrix}$
}
\label{equation:f_k_local}
\end{equation}

\begin{equation}
Q = 
\begin{bmatrix}
  \frac{\Delta t^4}{4}\sigma^2_{a_x} & 0 & \frac{\Delta t^3}{2}\sigma^2_{a_x} & 0 & 0 \\
  0 & \frac{\Delta t^4}{4}\sigma^2_{a_y} & \frac{\Delta t^3}{2}\sigma^2_{a_y} & 0 & 0 \\
  \frac{\Delta t^3}{2}\sigma^2_{a_x} & \frac{\Delta t^3}{2}\sigma^2_{a_y} & \Delta t^2\sigma^2_{a} & 0 & 0 \\
  0 & 0 & 0 & \Delta t^2\sigma^2_{\psi} & 0 \\
  0 & 0 & 0 & 0 & \Delta t^2\sigma^2_{\dot{\psi}} \\
\end{bmatrix}
\label{equation:q_local}
\end{equation}

\begin{equation}
z_{k} = \mu_{obs}
\label{equation:z_k_local}
\end{equation}

\begin{equation}
R_{k} = \Sigma_{obs}
\label{equation:r_k_local}
\end{equation}

\begin{equation}
H_{k} = \begin{bmatrix}
  1 & 0 & 0 & 0 & 0 \\
  0 & 1 & 0 & 0 & 0 \\
\end{bmatrix}
\label{equation:h_k_local}
\end{equation}

\section{Global Fusion Design}
\begin{figure}[ht]
  \centering
    \includegraphics[width=.45\textwidth]{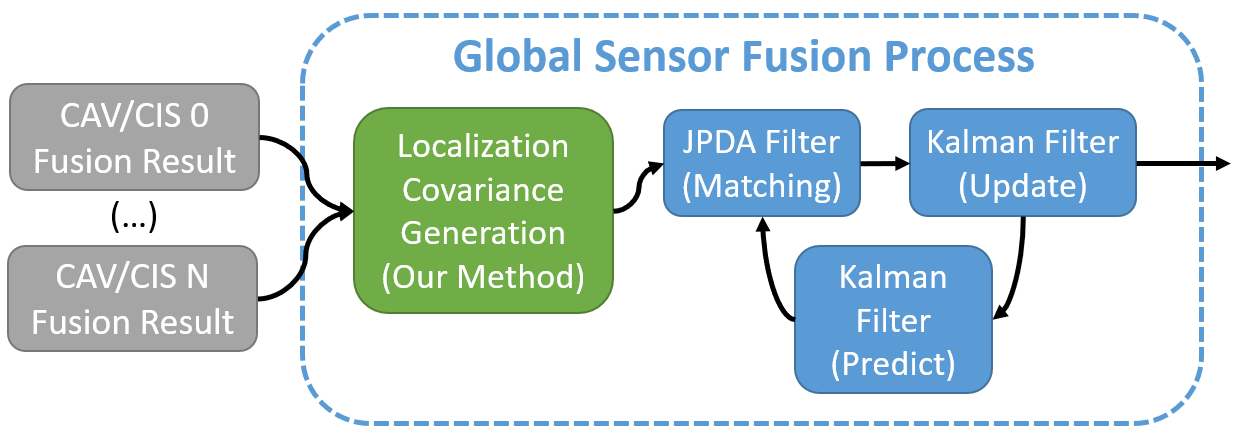}
    \caption{Our method for global sensor fusion is very similar to the local fusion, however it has an additional stage to generate the localization covariance matrix from the key predictor term velocity and combines that with the covariance generated from the local fusion.}
    \label{fig:global_fusion}
\end{figure}

A global cooperative sensor fusion is applied to an area larger than an ego vehicle can see on its own, e.g. the area around a traffic light, an entire city, etc. Our global sensor fusion shown in figure \ref{fig:global_fusion} is done similarly to our local fusion. However, instead of receiving observations from a set of sensors that are hard mounted on a vehicle body, we are working with a the local fusion outputs from a set of $n$ sensor platforms which may or may not be moving. These fused sensor outputs can be treated almost the same way as the single sensor pipelines, except for one major difference; the global sensor fusion must deal with localization error.

\subsection{Parameterized Localization Error Model}
Localization error must be accounted for when performing a global cooperative fusion of sensor values. Unlike on a sensor platform where the transformations between the sensors can be considered stationary because the sensors are rigidly mounted, the transformation between more than one moving sensor platform is not rigid and thus must be measured W.R.T. some shared coordinate system. Each sensor platform is assumed to have a localizer unless the platform is stationary like a CIS. For this stationary case, the CIS platform is considered to have a stationary measurement of location along with some error estimation of that measurement which will be treated the same as a localizer. A localizer regardless of type (SLAM, GPS, etc.) is assumed to report the current position of the sensor platform $sp$ with respect to a mutual global coordinate system (e.g. latitude and longitude). Our global fusion coordinate system can be thought of as a standard x,y coordinate system where each sensor platform position is reported as $<x_{sp}, y_{sp}>$. This localization needs an associated error as no localizer is perfect. As discussed in the error characterization section, we found an accurate predictor for the lateral error $\sigma_{lateral}$ and longitudinal error $\sigma_{longitudinal}$ component of localization to be measured velocity $v_{sp}$. We fit a polynomial to the longitudinal and lateral localization error using sensor platform velocity $v_{sp}$ as a predictor in equations \ref{equation:longitudinal} and \ref{equation:lateral_e} respectively. $\eta_k$ and $\gamma_k$ are the polynomial expansion coefficients for the longitudinal and lateral variance respectively that will be solved for.

\begin{equation}
\sigma_{longitudinal} = \sum_{k=0}^{n} \eta_k * v_{sp} 
\label{equation:longitudinal}
\end{equation}

\begin{equation}
\sigma_{lateral} = \sum_{k=0}^{n} \gamma_k * v_{sp}
\label{equation:lateral_e}
\end{equation}

Localization error can be represented as a 2x2 matrix, with the longitudinal $\sigma_{longitudinal}$ and lateral $\sigma_{lateral}$ error components estimated using the measured velocity the vehicle is traveling. This error can then be rotated based on the direction $\theta_{sp}$ the vehicle is traveling. These calculations are shown in equation \ref{equation:covariance_localization}. Measured location is converted to the global coordinate system using equation \ref{equation:vehicle_transform_global} with the sensor platform position $<x_{sp}, y_{sp}>$ and the position relative to the sensor platform that was generated by the local EKF state $<\hat{x}_{k|k}[0], \hat{x}_{k|k}[1]>$ (or $<x,y>$).

\begin{equation}
\mu_{sp\_obs} = \big[\begin{smallmatrix}
  x_{sp} \\
  y_{sp}
\end{smallmatrix}\big] + \big[\begin{smallmatrix}
  \hat{x}_{k|k}[0] \\
  \hat{x}_{k|k}[1]
\end{smallmatrix}\big]
\label{equation:vehicle_transform_global}
\end{equation}

\begin{equation}
\Sigma_{sp} = 
\big[\begin{smallmatrix}
  cos(\theta_{sp}) & sin(\theta_{sp}) \\
  -sin(\theta_{sp}) & cos(\theta_{sp})
\end{smallmatrix}\big]
\big[\begin{smallmatrix}
  \sigma_{longitudinal} & 0 \\
  0 & \sigma_{lateral}
\end{smallmatrix}\big]
\big[\begin{smallmatrix}
  cos(\theta_{sp}) & sin(\theta_{sp}) \\
  -sin(\theta_{sp}) & cos(\theta_{sp})
\end{smallmatrix}\big]^\intercal
\label{equation:covariance_localization}
\end{equation}

With the covariance of the localization known, we combine that error with the local fusion EKF error from each platform. The EKF outputs the covariance $P_{k|k}$ but it must be rotated to the world coordinate system using equation \ref{equation:covariance_local_to_global}. This covariance from $P_{k|k}$ already contains the perception error that was factored in from the local fusion. We can sum the $P_{k|k}$ covariance with the localization covariance to get an approximate covariance for that sensor fusion track combined with the expected covariance of the sensor platform localization using equation \ref{equation:gaussian_union} \cite{allig2020unequal}. This is not an exact covariance merger as it does clip some of the data that would be there with a higher-order representation. However, this is viewed as a worthwhile trade-off since our sensor fusion algorithms expects a 2x2 covariance input, thus a higher-order representation would not be usable.

\begin{equation}
\resizebox{\linewidth}{!}{$
\Sigma_{obs\_world} = 
\big[\begin{smallmatrix}
  cos(\theta_{sp}) & sin(\theta_{sp}) \\
  -sin(\theta_{sp}) & cos(\theta_{sp})
\end{smallmatrix}\big]
\big[\begin{smallmatrix}
  P_{k|k_{[0][0]}} & P_{k|k_{[0][1]}} \\
  P_{k|k_{[1][0]}} & P_{k|k_{[1][1]}}
\end{smallmatrix}\big]
\big[\begin{smallmatrix}
  cos(\theta_{sp}) & sin(\theta_{sp}) \\
  -sin(\theta_{sp}) & cos(\theta_{sp})
\end{smallmatrix}\big]^\intercal
$}
\label{equation:covariance_local_to_global}
\end{equation}

\begin{equation}
\Sigma_{sp\_obs} = [\Sigma_{sp}^\intercal + \Sigma_{obs\_world}^\intercal]^\intercal
\label{equation:gaussian_union}
\end{equation}

\subsection{Global Fusion Incoming Message Packet Design}
We assume global sensor fusion computations to happen on a Road Side Unit (RSU) which is a computation platform located near the area of interest that can communicate with CAVs and CISs in the area. Data is packetized to be sent to the RSU and includes sensor platform localization $\mu_{sp\_obs}$, estimated localization covariance for $\Sigma_{sp\_obs}$ generated in equations \ref{equation:vehicle_transform_global} and \ref{equation:gaussian_union} respectively as well as the vehicles own reported localization $\mu_{sp}$ and estimated covariance $\Sigma_{sp}$.

\subsection{Global Fusion Joint Probability Data Association Filter}
The sensor platform JPDA Filter shares the same design as the one used locally on the sensor platforms. The only difference is now the sensing platforms report their observations W.R.T. a global coordinates using equation \ref{equation:vehicle_transform_global} and \ref{equation:gaussian_union}. This uses the same $\mu$ and $\sigma$ form as the local fusion so no changes to the JPDA Filter design are necessary.

\subsection{Global Fusion EKF Model}
The EKF for the global fusion is nearly identical to the EKF used in the local EKF. The state $x_k$, update matrix $F_k$, $Q$, and $H_k$ for the global fusion EFK is the same as those used by the local fusion EKF in equations \ref{equation:x_k_local}, \ref{equation:f_k_local}, \ref{equation:q_local}, and \ref{equation:h_k_local} respectively. We change the $z_k$ and $R_k$ values as shown in equation \ref{equation:glob_z_k} and \ref{equation:glob_r_k} respectively. Just like the local fusion, the update rate is constant and the global fusion EKF has a predict stage and then up to $s$ update stages dictated by the number of sensors packages (CAVs and CISs) there are reporting observations that are associated to the same track by the JPDA Filter.

\begin{equation}
z_{k_{sp}} = \mu_{sp\_obs}
\label{equation:glob_z_k}
\end{equation}

\begin{equation}
R_{k_{sp}} = \Sigma_{sp\_obs}
\label{equation:glob_r_k}
\end{equation}

\section{Cooperative Fusion Experimental Results}
Our 1/10 scale model experiment utilizes the map sizes and vehicle counts listed in table \ref{table:maps}. We ran ten minute tests of the four setups depicted five times each. Data was recorded during all 20 tests so that the test set would be the same as we ran the different fusion methods. CIS sensor data is simply removed from the data-set when it not needed so that makes 8 total scenarios or 40 tests. The Optitrack system was calibrated to an RMSE of $0.00088m$. We compare two error models: 1) our parameterized error model discussed before and 2) a typical fixed error model as a baseline. For the fixed error model, we use the mean error that we gathered using sensor classification for each sensing pipeline and the localization pipeline. The equations used for our parameterized error model can be found in table \ref{table:sensing_error}. We test two different filters, the EKF outlined above from Farag \textit{et al.} as well as the UKF proposed in the paper by Garcia \textit{et al.}, which we refer to as ``Farag EKF'' and ``Garcia UKF'' respectively \cite{farag2021kalman, garcia2017sensor}.

\begin{table}[]
\caption{\textbf{1/10 Scale Sensing Expected Variance Equations}}
\label{table:sensing_error}
\begin{tabular}{llll}
 \textbf{Error Type} & \textbf{Variable} & \textbf{Parameterized} & \textbf{Fixed} \\
 \textit{(Direction, Sensor)}  &  & \textit{(Our Method)} & \textit{(Mean)} \\
 Distal, Camera & $\sigma_{distal}$ & $0.0517*d + 0.0126$ & 0.0881 \\
 Perpendicular, Camera & $\sigma_{perp}$ & $0.0117*d + 0.023$ & 0.0401 \\
 Distal, LIDAR & $\sigma_{distal}$ & $0.0165*d + 0.0607$ & 0.0848 \\
 Perpendicular, LIDAR & $\sigma_{perp}$ & $0.0097*d + 0.0361$ & 0.0503 \\
 Longitudinal, Localizer & $\sigma_{longitudinal}$ & $0.0782*v + 0.0428$ & 0.0663 \\
 Lateral, Localizer & $\sigma_{lateral}$ & $0.0841*v + 0.0241$ & 0.0493 \\
\end{tabular}
\end{table}

\begin{figure}[ht]
  \centering
    \includegraphics[width=.48\textwidth]{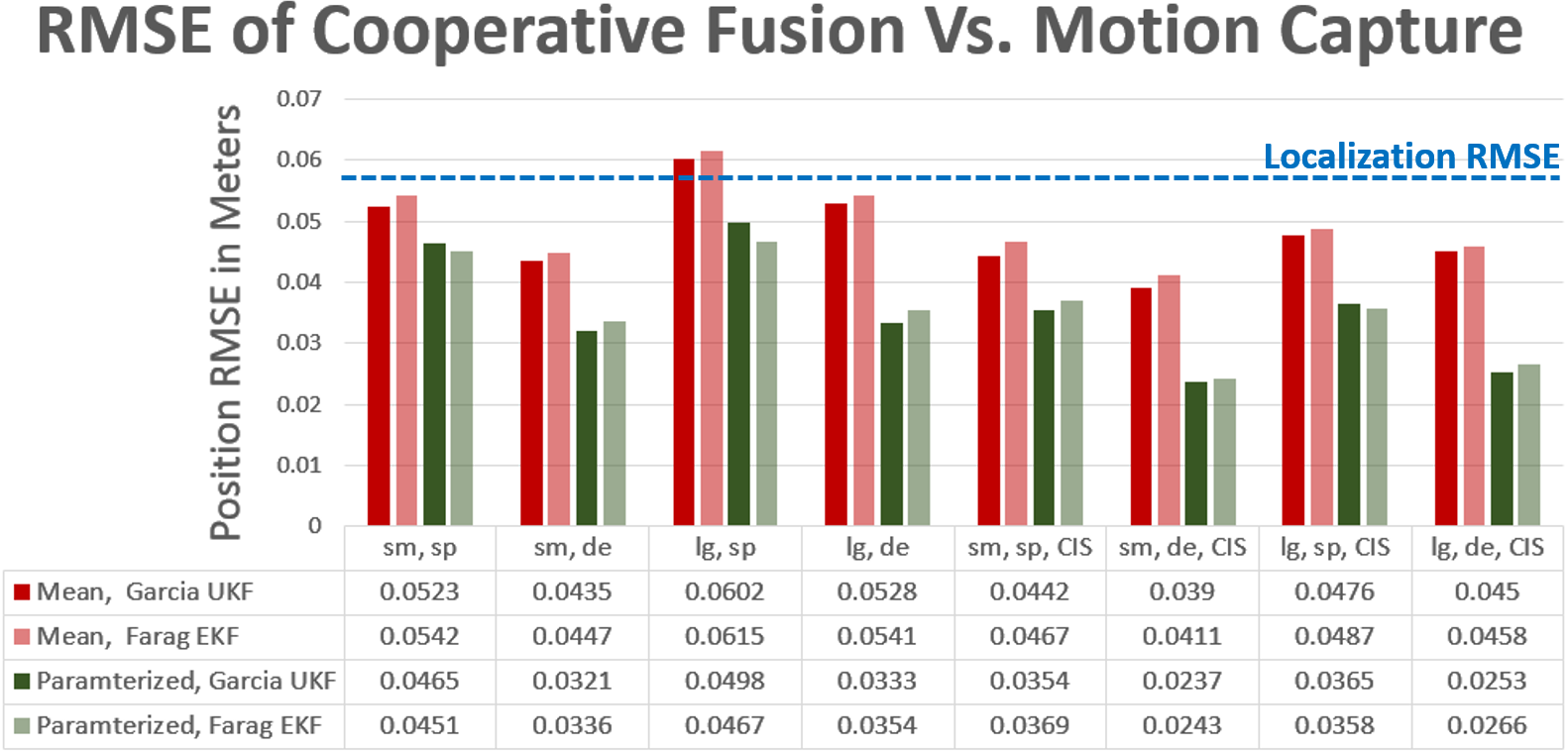}
    \caption{Results of the cooperative fusion method using a fixed vs. parameterized error model. ``Garcia UKF'' is from Garcia \textit{et al.}\cite{garcia2017sensor} while ``Farag EKF'' is described in the prior section \cite{farag2021kalman}. RMSE for the localization of the CAVs alone is shown as a blue line - Cooperative Fusion RMSE needs to be less than this line for results to be considered more accurate than localization data alone.}
    \label{fig:results}
\end{figure}

\subsection{A Fixed Error Approach can Result in Less Accurate CAV Positions than Onboard Localization Alone}
RMSE localization error for the M1M1 LIDAR is 0.0576 meters. Therefore in order for the cooperative sensor fusion to return data that is useful, reported RMSE vs. the Optitrack motion capture system must be less than 0.0576m. A major result of note is that when using a fixed model, some of the tests get an overall RMSE near 0.0576m. This can be seen very clearly in the large sparse no CIS scenario ''ls,sp" where the RMSE is 0.0602m and 0.0615m, which is worse than the localizer alone for both methods. Our parameterized error model allows the filter to get an RMSE value that is equal or less RMSE than the localisation in all cases.

\subsection{Connected Infrastructure Sensors Are Useful Due to Lack of Localization Error}
In general the larger map with the sparse setting (i.e. fewer sensors) had a higher RMSE due to the larger distances and lack of sensing. The small dense maps on the other hand have results with the lowest RMSE values. When CIS sensors are added, regardless of the fixed or parameterized model, RMSE decreases because the position of these CIS sensors is well known and does not change thus there is very little error introduced by the localization step. This makes CIS sensors very useful compared to the CAVs which have an average localization error of 0.0576m.

\subsection{Our Parameterized Sensor Fusion Approach is More Accurate than a Fixed Approach}
The most significant finding of these results is that our parameterized error model outperforms the fixed error model in every scenario. In the best case, our parameterized error model along with our proposed EKF achieves an RMSE of 0.0237m which is 2.43x better than localization was able to achieve alone and it was 1.7x better than the fixed error model in the same scenario. The results also show that our parameterized error model decreases RMSE for both our EKF and Garcia UKF models showing that it is useful for more than just one specific EKF. 

\section{Conclusion}
In this paper, we propose a parameterized sensing and localization error model for use in connected autonomous vehicles to improve cooperative sensor fusion. We performed an analysis of scale 1/10 model autonomous vehicles with scale-accurate sensors to fit the model. To our knowledge, this is the first time that a comprehensive error classification of an autonomous vehicle sensing suite has been performed against a baseline sensing system that is accurate enough to evaluate the results. Results of our parameterized error model are integrated with a tiered, high level cooperative sensor fusion pipeline using an EKF. Our results show an average improvement of 1.42x in RMSE versus a typical fixed error model on our 1/10 scale test-bed. In the future, we hope to perform the same analysis on full-size autonomous vehicles to prove it scales with size as well as analyze the possible effect of weather conditions, lighting, city density, etc. as predictors in a less controlled environment.

\addtolength{\textheight}{-12cm}   









\bibliographystyle{IEEEtranS}
\bibliography{IEEEabrv,root}

\end{document}